\pdfoutput=1

\documentclass[11pt]{article}

\usepackage[]{acl}

\usepackage{times}
\usepackage{latexsym}

\usepackage[T1]{fontenc}

\usepackage[utf8]{inputenc}

\usepackage{microtype}

\usepackage{inconsolata}

\usepackage{graphicx}

\usepackage{times}
\usepackage{latexsym}
\usepackage{enumitem}
\usepackage{supertabular}
\usepackage{amsmath}

\usepackage[T1]{fontenc}

\usepackage{scalerel,graphicx,xparse}

\NewDocumentCommand\emojipatient{}{\raisebox{-0.2ex}{\scalebox{1.5}{\scalerel*{\includegraphics{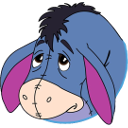}}{0}}}}

\usepackage{url}

\usepackage[nopar]{lipsum}
\usepackage{amsmath}
\usepackage{amsmath, amssymb, amsfonts}

\usepackage{graphics}
\usepackage{CJKutf8}
\usepackage{tabularx}
\usepackage{booktabs}
\usepackage{multicol}
\usepackage{multirow}
\usepackage{mathtools}
\usepackage{graphicx} 
\usepackage{ragged2e} 
\usepackage{makecell}
\usepackage{tablefootnote}
\usepackage{array}
\usepackage{makecell}
\usepackage{cleveref}
\usepackage{algorithm2e}
\usepackage{enumitem}
\usepackage{subcaption}
\usepackage[skip=1ex]{caption}
\usepackage[noend]{algpseudocode} 
\usepackage{hyperref}
\newcommand\rurl[1]{\href{http://#1}{\nolinkurl{#1}}}
\usepackage{pifont}
\usepackage[normalem]{ulem}

\usepackage{tabularx}
\usepackage{xurl}

\usepackage{circledsteps} 
\usepackage{fontawesome5}

\title{\emojipatient Eeyore: Realistic Depression Simulation via \\Supervised and Preference Optimization}

\author{
 Siyang Liu$^{1}$\textsuperscript{\href{mailto:lsiyang@umich.edu}{\faEnvelope}},
 \textbf{Bianca Brie\textsuperscript{1}},
 \textbf{Wenda Li\textsuperscript{1}},
 \\
 \textbf{Laura Biester\textsuperscript{2}},
 \textbf{Andrew Lee\textsuperscript{3}},
 \textbf{James W Pennebaker\textsuperscript{4}},
 \textbf{Rada Mihalcea$^{1}$\textsuperscript{\href{mailto:mihalcea@umich.edu}{\faEnvelope}}}
\\
\\
 \textsuperscript{1}The LIT Group, Department of Computer Science and Engineering, \\University of Michigan, Ann Arbor,
 \\
 \textsuperscript{2}Middlebury College,
 \textsuperscript{3}Harvard University,\\
 \textsuperscript{4}Department of Psychology, University of Texas at Austin, Austin
\\
}

\begin{document}
\maketitle
\begin{abstract}
Large Language Models (LLMs) have been previously explored for mental healthcare training and therapy client simulation, but they still fall short in authentically capturing diverse client traits and psychological conditions. We introduce \textbf{Eeyore}, an 8B model optimized for realistic depression simulation through a structured alignment framework, incorporating expert input at every stage.
First, we systematically curate real-world depression-related conversations, extracting depressive traits to guide data filtering and psychological profile construction, and use this dataset to instruction-tune Eeyore for profile adherence. Next, to further enhance realism, Eeyore undergoes iterative preference optimization---first leveraging model-generated preferences and then calibrating with a small set of expert-annotated preferences.
Throughout the entire pipeline, we actively collaborate with domain experts, developing interactive interfaces to validate trait extraction and iteratively refine structured psychological profiles for clinically meaningful role-play customization.
Despite its smaller model size, the Eeyore depression simulation outperforms GPT-4o with SOTA prompting strategies, both in linguistic authenticity and profile adherence.

\end{abstract}

\section{Introduction}
Psychological science, like other scientific domains such as chemistry, physics, medicine, and neuroscience \cite{thirunavukarasu2023large, demszky2023using, boiko2023autonomous, birhane2023science}, has increasingly recognized the transformative power of large language models (LLMs) to advance the field \cite{demszky2023using}. Recent studies have shown that LLMs can support psychology in areas like measurement \cite{wang2024towards, wang2024explainable}, experimentation \cite{argyle2023out}, and clinical practice \cite{wang2024patient}. In particular, leveraging the role-playing capabilities of LLMs to simulate therapy-related roles, for example, a client with ongoing depression, has shown promise in helping novice counselors or psychiatrists practice their clinical skills \cite{wang2024patient, louie2024roleplay}.

However, despite their promise, existing LLM-driven simulations face limitations that hinder their adoption in professional clinical training. 
Current approaches rely heavily on prompt engineering \cite{qiu2024interactive, wang2024patient, louie2024roleplay, wang2024towards, qiu2024interactive}, which cannot overcome the inherent biases and structural constraints of general-purpose LLMs \cite{haltaufderheide2024ethics}.
Recent studies have raised concerns about the validity of using LLMs for clinical training, particularly regarding their inability to authentically represent patient experiences and their tendency to generate overly sanitized or misleading responses \cite{feigerlova2025systematic, zidoun2024artificial, gabriel-etal-2024-ai, zhui2024ethical, haltaufderheide2024ethics,wang2024patient}.
These concerns highlight the need for a structured approach that moves beyond generic prompting strategies.

\begin{figure*}[t]
    \centering
    \includegraphics[width=0.8\linewidth]{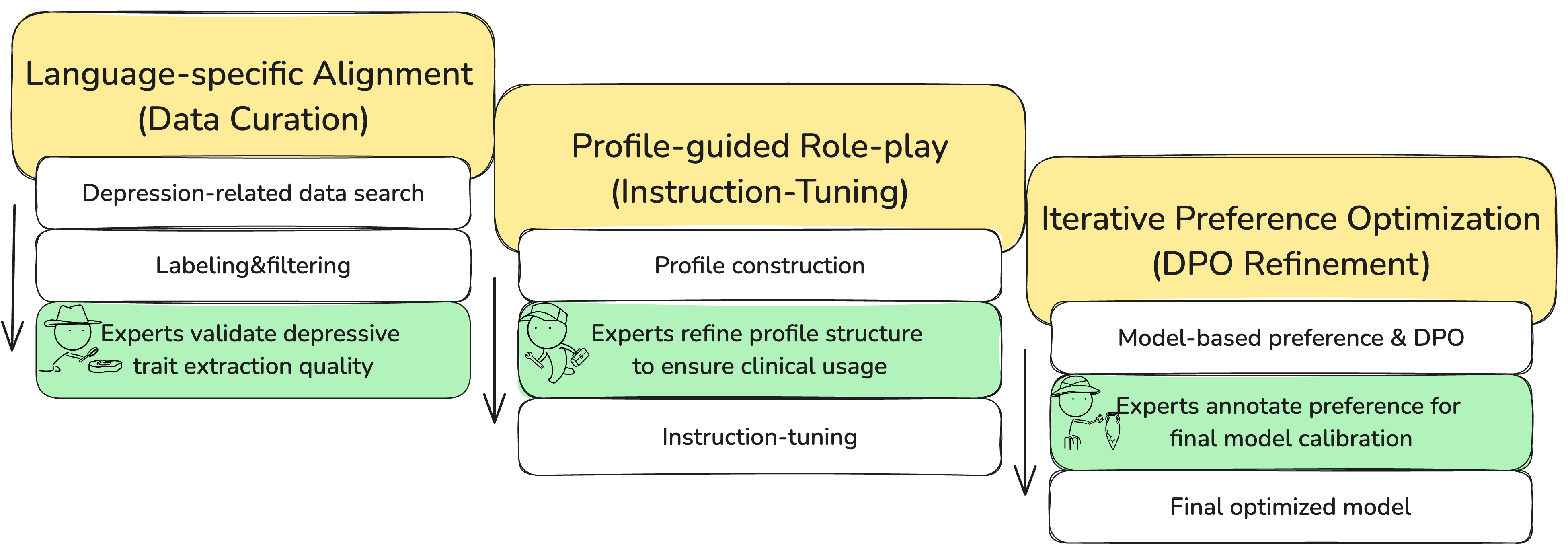}
    \caption{The alignment pipeline for optimizing LLMs to simulate individuals with depression in clinical training. Expert involvement is highlighted in green. \textit{Icons by \citet{kudinovs2024icon}.}}
    \vspace{-3mm} 
    \label{fig:pipeline}
\end{figure*}

In this work, we develop a \textbf{structured alignment framework} to optimize LLMs for capturing the language, cognitive patterns, and experiential traits of individuals with depression in clinical training scenarios. As outlined in Figure~\ref{fig:pipeline}, our framework integrates \textbf{three specialized alignment endeavors} in a sequential pipeline, incorporating expert feedback at each stage. The three key innovations in our framework are:
\vspace{-3mm}
\paragraph{Language-specific Alignment.}
    As noted by \citet{haltaufderheide2024ethics}, biases in training data can undermine the authenticity of simulated patient interactions. 
    General-purpose LLMs (e.g., GPT-4) are optimized to prioritize being positive, supportive, and safe, which creates an inherent ceiling on their ability to simulate depressive speech patterns (e.g., self-harm ideation, or cognitive distortions), even with carefully crafted prompting.
    To bridge this gap, we conduct an extensive search across public resources and datasets to find \textbf{real-world depression-related conversations}, which are often buried within broader corpora. We leverage a combination of GPT-4o \footnote{All mentions of GPT-4o in this paper refer to GPT-4o (2024-08-06) \cite{openai2024gpt4o}} labeling, existing annotated data, and careful filtering techniques to systematically mine, extract, and rebalance data. This process curates 3,042 high-quality conversations, ensuring comprehensive coverage of diverse depressive traits and realistic conversational settings. This resource serves as a solid data foundation for modeling authentic depressive language and cognitive patterns.

\paragraph{Profile-Guided Role-Playing via Instruction-Tuning.} 
    Depression manifests uniquely in each individual, and clinical training requires exposure to varied cases of depression for customized practice. To achieve this, we structure each conversation in our dataset with a corresponding psychological profile that encodes important information about depressive traits. 
    These profiles undergo \textbf{iterative refinement through expert critiques} to ensure clinical accuracy and relevance.
    We instruction-tune an LLM using system prompts that specify the client’s profile and conversation context, allowing it to role-play with consistency and realism across different depressive manifestations. This approach lets practitioners engage with a broad spectrum of depressive profiles, mirroring real-world variations in symptoms and experiences.

\paragraph{Iterative Preference Optimization.}
    While instruction-tuning improves adherence to psychological profiles, further refinement is needed to align the model’s outputs with expert expectations.  Given the high cost of expert annotation, we adopt a two-stage direct preference optimization (DPO) \cite{rafailov2023direct} process:
\vspace{-2mm}
\begin{itemize}[leftmargin=0.3cm, itemsep=-0.8pt]
    \item Stage 1 (Model-Based Preference Generation). We generate 1,933 preference data samples using a model-based verifier, employing a novel sampling method that adds a small amount of noise to psychological profiles to produce highly contrasting negative responses. This overcomes the model’s tendency to generate only subtle deviations in sampling, further facilitating models to learn clear distinctions between preferred (fully aligned) and less preferred (slightly deviated) responses. The model is then trained on this preference data via DPO.
    The model trains on this preference data via DPO.
    \item Stage 2 (Expert Preference Calibration). We collect 250 human-annotated preference labels from expert counselors to fine-tune the DPO model as a final calibration. This calibration step ensures that the model \textbf{aligns with expert expectations while keeping annotation costs minimal.}
\end{itemize}
Through comprehensive evaluations, Eeyore is found to outperform state-of-the-art baselines based on GPT-4o in both linguistic authenticity and profile adherence. Expert evaluations highlight Eeyore’s ability to produce natural, emotionally nuanced responses while adhering to assigned psychological profiles.
Our findings demonstrate that structured optimization beyond prompt engineering is crucial for achieving more clinically satisfactory LLM-driven simulations.
We invested in interactive interfaces for online testing, hoping to move LLM-based mental health training beyond labs and encourage expert adoption with confidence.

\section{Related Work}
\paragraph{LLM-Based Patient Simulation in Mental Health.} 
Recent work has explored using LLMs to simulate therapy clients for clinician training \cite{wang2024patient, louie2024roleplay}. Early approaches relied on generic LLM prompting \cite{qiu2024interactive}, but concerns about clinical validity and ethical risks \cite{haltaufderheide2024ethics, zidoun2024artificial} have led to structured modeling efforts. Patient-$\psi$ \cite{wang2024patient} integrates cognitive modeling from clinical frameworks to enhance realism, while Roleplay-doh \cite{louie2024roleplay} applies principle-adherence prompting to improve consistency. However, these methods struggle with generating nuanced, profile-consistent responses, highlighting the need for systematic alignment strategies.

\paragraph{Preference Optimization for Alignment.}
Optimizing LLMs with human preference data has been widely studied \cite{christiano2017deep}, with Direct Preference Optimization (DPO) emerging as an efficient alternative to reinforcement learning \cite{rafailov2023direct}. While DPO has been applied in general chatbot alignment and some scientific domains \cite{cheng2024decomposed,savage2024fine}, its use in simulation for clinical psychology practice remains underexplored. Recent methods propose augmenting preference data through automated techniques \cite{pi2024strengthening, lu2024step}, which aligns with our approach of leveraging model-based augmentation to enhance preference learning for profile-guided mental health simulations.

\section{Methodology}

Our framework, illustrated in Figure~\ref{fig:pipeline}, consists of three stages: (1) Language-Specific Alignment, where we curate a dataset of depression-related conversations with structured psychological profiles; (2) Profile-Guided Role-Playing, where we instruction-tune the model for realistic profile adherence; and (3) Iterative Preference Optimization, where we refine the model via model-generated and expert-annotated preferences.

\subsection{Language-specific Alignment}
\paragraph{Depression-related Data Search.} 

We collect depression-related conversations from publicly available sources, including mental health forums, clinical transcripts, and academic datasets. The selected datasets include: (1) \textbf{RED} \cite{welivita-etal-2023-empathetic}: threads from subreddits r/depression and r/depressed, structured as dialogues.
(2) \textbf{HOPE} \cite{10.1145/3488560.3498509}: transcripts from publicly available pre-recorded counseling videos on YouTube.
(3) \textbf{ESC} \cite{liu-etal-2021-towards}: a dataset of crowdsourced emotional support conversations.
(4) \textbf{AnnoMI-Full} \cite{9746035}: transcripts of therapy sessions demonstrating motivational interviewing skills.
These datasets qualify for our study based on the following criteria:
(i) All conversations must be produced by humans instead of AI-synthesized.
(ii) They must feature multi-turn conversations.
(iii) They are from publicly available sources.
(iv) They are relevant to mental health, and at least one participant is likely experiencing emotional distress, though not necessarily diagnosed with depression.
After gathering these datasets, we process 5,618 conversations for further labeling and filtering.

\paragraph{Labeling \& Filtering.} 

To determine whether a conversation is depression-related, we apply the following criteria: the conversation exhibits at least one core depression feature, such as loss of interest in activities or depressed mood.
We do not require participants to have a formal depression diagnosis, as such data is typically unavailable in public sources.
We employ GPT-4o for automated classification. For HOPE and AnnoMI-Full, GPT-4o identifies depression-related conversations.
For ESC, we use pre-existing labels to extract relevant conversations.
For RED, we adopt all conversations as depression-related.
To ensure a balanced dataset, we analyze depressive traits across conversations using GPT-4o-based extraction. 
Our depressive traits are structured according to psychological profiles (introduced in Section \ref{subsec:role-play}).

After extraction, we observe significant imbalances in symptom severity and demographic attributes. For example, moderate and severe depression cases are overrepresented compared to minimal and mild cases, which will introduce role-play bias if directly tuning on them.
To alleviate bias, we filter and rebalance the dataset, ultimately selecting 3,042 conversations \footnote{Eeyore and all annotated data will be open-sourced at \url{anonymous.github.com}.}. The final trait distribution is presented in Tables \ref{tab:dataset_traits_1} and \ref{tab:dataset_traits_2} of the Appendix.

\paragraph{Expert Review.} \label{subsec:languagealign}
To evaluate the accuracy of depression trait extraction, which is crucial for both data rebalancing and psychological profile construction, we recruit six experts specializing in clinical psychology, counseling psychology, social psychology, or social work.\footnote{All experts in this study, including those in expert review, profile refinement, and preference annotation, were recruited via Prolific (\url{https://www.prolific.com}), a widely used platform for academic research.} Each expert reviews three conversation scripts alongside their extracted psychological traits.
Each conversation contains approximately 20 extracted traits, such as age, specific symptoms, and cognitive distortions. Experts assessed whether each trait accurately reflects the conversation.
Expert responses are categorized as: "Yes, it is directly reflected in the conversation", "Yes, it is a reasonable inference, though not directly stated"
, or "No, it does not accurately reflect the conversation".
Overall, 85.2\% of extracted traits were verified as accurate extraction.
Among these, 57.6\% acknowledge indirect but reasonable inferences made by the model.


\subsection{Profile-Guided Role-Play} \label{subsec:role-play}

\paragraph{Psychological Profile Construction.} 
The psychological profile serves as a structured representation of the client in the conversation. Its design requires cross-disciplinary collaboration between AI researchers and mental health professionals. We first develop a preliminary profile, considering what information can be realistically extracted from conversations and how an initial model can be trained to allow experts to refine the profile within their context of use.

Each profile consists of three parts: \textit{demographics}, including general information (e.g. gender, occupation); \textit{situational context}, which captures distress-related situations and attitudes toward seeking support; and \textit{depression-related manifestations}, which describe symptoms and cognitive patterns (see Table~\ref{tab:original_profile} for the original design and modifications).
Among these, \textit{depression-related manifestations} are the most clinically relevant. We review foundational psychological literature and structure it as follows: \textit{Depression symptoms} are extracted from \textit{DSM-V} \cite{edition2013diagnostic}, where 18 related symptoms are categorized as \textit{not exhibited, mild, moderate, or severe}. \textit{Cognitive distortions} are adapted from Beck’s theory \cite{clak1999scientific, beck2009depression}, identifying 5 thought patterns labeled as \textit{exhibited} or \textit{not exhibited}. \textit{Functional impairments} \cite{ustun2010measuring} were initially included but later removed following expert feedback. \textit{Overall depression severity} follows a four-level categorization (\textit{minimal, mild, moderate, severe}), inspired by \textit{PHQ-9} \cite{kroenke2001phq} and \textit{DBI} \cite{beckbeck}.

We extract structured profiles from all conversations using GPT-4o and use them to train an initial instruction-tuned model that role-plays clients based on these profiles.
Experts then interact and evaluate this model, as described in the following paragraph.
We refine the profiles and instruction-tuning dataset based on their feedback.

\begin{figure}[thb]
    \centering
    \includegraphics[width=\linewidth]{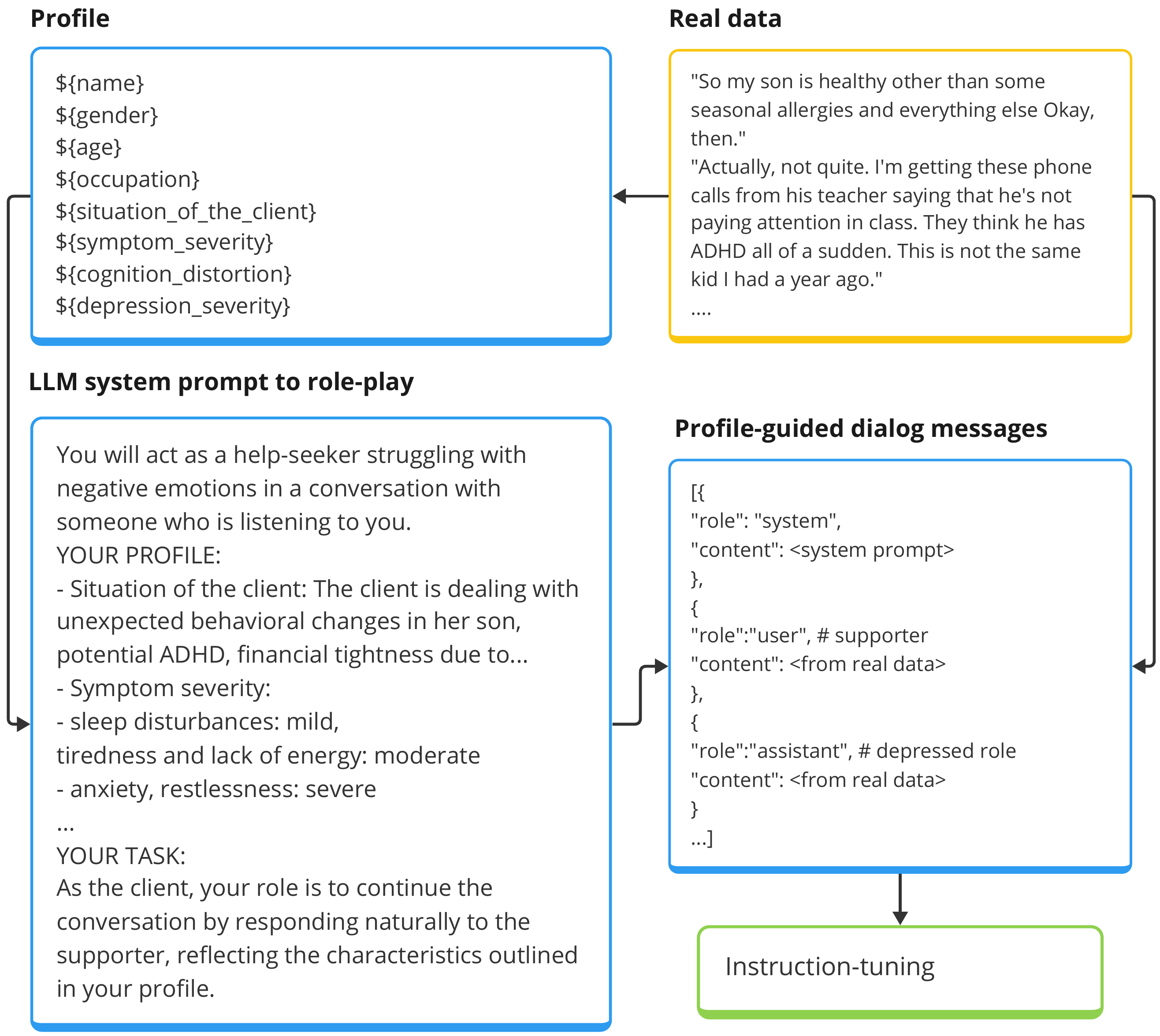}
    \caption{\label{fig:data_construct}Pipeline to input data for instruction-tuning.}
    \vspace{-3mm}
    \label{fig:instr}
\end{figure}

\paragraph{Expert Profile Refinement.} 
To refine client profiles, we conduct a pilot study (see survey details and interaction interface in Appendix~\ref{appx:refine}) with ten experts. 
They interact with the model trained on the preliminary version of the profile by customizing attributes and engaging in conversations. This interactive evaluation highlights areas for improvement while validating the overall structure. 

Among the profile attributes in the initial design, 80\% receive expert approval, while some items are reported as ambiguous or redundant. Based on expert feedback, we remove \textit{unwillingness to express feelings} (redundant), \textit{emotional fluctuation} (ambiguous), and \textit{functional impairment} (overlaps with specific symptoms). Additionally, we add \textit{marital status}, \textit{counseling history}, \textit{suicidal ideation severity}, and \textit{homicidal ideation severity}, as they provide critical contextual relevance. Two additional suggestions, \textit{period of depression} and \textit{current treatment}, are not included as they cannot be reliably extracted from available conversations.
To accommodate the need for \textit{counseling history}, we construct multi-session interactions by segmenting lengthy conversations and summarizing prior sessions. This enables 453 out of 3,042 data points in the dataset to now include counseling history. The revised profile is exemplified in Figure~\ref{fig:client_profile}, and the refined dataset is used for re-extraction and instruction-tuning. See Section~\ref{subsec:languagealign} for the extraction accuracy.

\paragraph{Instruction-Tuning.}

Figure~\ref{fig:instr} shows our procedure to convert our data into an instruction-tuning format. 
After integrating expert feedback, we extract updated profiles and reconstruct the instruction-tuning dataset. The structured profile is embedded in the system prompt, while the assistant’s messages simulate the responses of a depressed client. The model is trained to predict the assistant’s utterances while treating system prompts and user messages as context. This ensures the model generation is consistent with the assigned profile, improving realism in role-play interactions.

\subsection{Iterative Preference Optimization}

\begin{figure}[thb]
    \centering
    \includegraphics[width=0.8\linewidth]{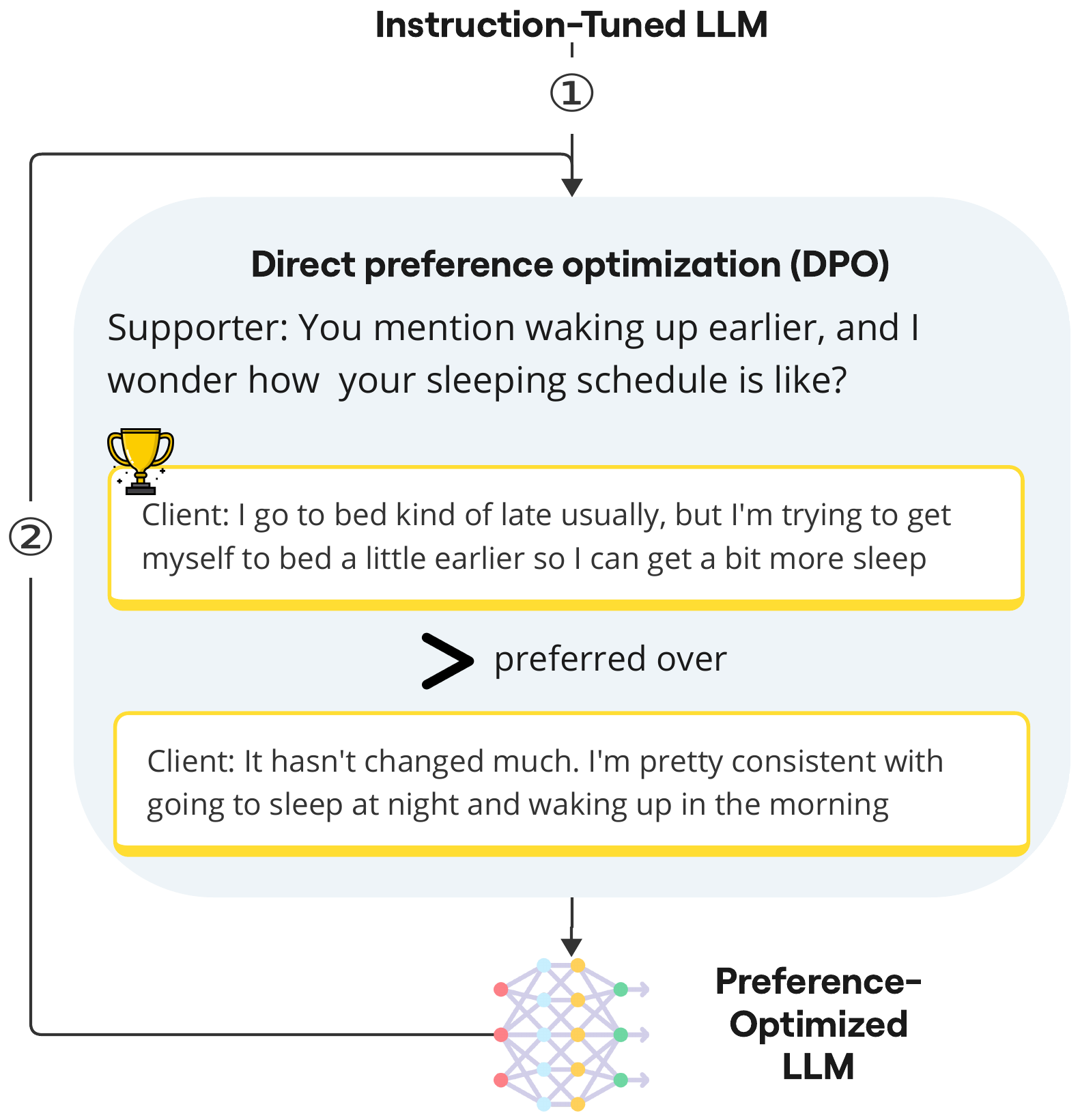}
    \caption{Overview of the two-stage Direct Preference Optimization process.  \Circled{1} optimizes a DPO model from the instruction-tuned model using model-based preference data. \Circled{2} refines the DPO model with expert-annotated preferences, producing the final preference-optimized model.}
    \label{fig:dpo}
\end{figure}

While instruction-tuning improves profile adherence, further refinement is required to align model outputs with expert expectations. We adopt a two-stage direct preference optimization (DPO) approach \cite{rafailov2023direct}, first leveraging model-generated preferences and then refining with expert annotations (see Figure~\ref{fig:dpo}).

\paragraph{Iterative DPO Training.} 

The DPO loss function optimizes the policy model $\pi_{\theta}$ relative to a reference model $\pi_{\text{ref}}$, enforcing preference alignment:

\small
\begin{align}
\mathcal{L}_{\text{DPO}}(\pi_{\theta}; \pi_{\text{ref}}) &= 
- \mathbb{E}_{(x, y_w, y_l) \sim \mathcal{D}} 
\Bigg[ \log \sigma \Big( 
\beta \log \frac{\pi_{\theta}(y_w \mid x)}{\pi_{\text{ref}}(y_w \mid x)} 
\Big) \nonumber \\
&\quad - \log \sigma \Big( 
\beta \log \frac{\pi_{\theta}(y_l \mid x)}{\pi_{\text{ref}}(y_l \mid x)}
\Big) \Bigg].
\end{align}
\normalsize
where $(x, y_w, y_l)$ represents the input prompt (in our case, a profile-guided dialog context), the preferred response, and the less preferred response, respectively. The model is trained to distinguish between responses while remaining aligned with the reference model.

As illustrated in Figure~\ref{fig:dpo}, the optimization process consists of two phases. 
In the first phase, we take the instruction-tuned model as $\pi_{\text{ref}}$, and optimize it using model-generated preference data, producing an intermediate DPO model as $\pi_{\theta}$.
 We further refine $\pi_{\theta}$ by using expert-annotated preferences, treating the previously optimized model as $\pi_{\text{ref}}$, and obtaining the final preference-optimized model $\pi_{\theta}$.

\paragraph{Model-based Preference Generation.} 
A classical approach to preference generation involves sampling two responses $(y_w, y_l) \sim \pi_{\text{ref}}(y \mid x)$ from the same source prompt $x$. However, in our case, this method is ineffective. 
Our instruction-tuned model already exhibits strong profile-following ability, making it difficult to generate clearly distinguishable good and bad responses from the same input.
To assess its adherence to psychological profiles, we go through an evaluation on more than 4,000 model-generated responses using a GPT-4o-based verifier to score whether the response aligns with the client profile.
On average, a response will comply with 96.0\% of the attributes in the corresponding profile.
However, only 31.7\% of responses fully match all attributes, suggesting that while the model performs well, it generally is not perfect and still generates subtle inconsistencies.

This observation makes standard preference generation ineffective, as most responses are either both good or only slightly flawed, making it difficult to establish clear preference distinctions. Inspired by prior work adopting automatic negative response collection \cite{pi2024strengthening,lu2024step}, we introduce a contrastive augmentation strategy that artificially amplifies response differences. Specifically, we apply profile noise augmentation, where we modify 30\% of a psychological profile’s attributes (e.g., changing a symptom’s severity from “severe” to “mild”). We then generate a response $y_n$ using the modified profile:

\small
\begin{equation}
    y_n \sim  \pi_{\text{ref}}(y \mid x_n), \quad y_o \sim  \pi_{\text{ref}}(y \mid x_o),
\end{equation}
\normalsize

where $x_n$ represents the noisy prompt, and $x_o$ is the original.

However, this introduces a risk: since $y_n$ is generated from a different input than $y_o$, it can theoretically not be a naturally likely response from the reference model’s original prompt distribution. To mitigate this, we apply two selection criteria:

\begin{enumerate}[leftmargin=15pt, itemsep=-0.8pt]
    \item \textbf{Profile Adherence Score Constraint}: The GPT-4o verifier assigns an adherence score $S(y \mid x)$ based on how well a response follows the given profile. We enforce:
    \small
    \begin{equation}
        S(y_o \mid x_o) > S(y_n \mid x_n),
    \end{equation}
    \normalsize
    ensuring that $y_o$ aligns better with its profile than $y_n$ does with its noisy profile.
    
    \item \textbf{Generation Probability Ratio Constraint}: We define the average token probability of a response $y$ under the original prompt $x_o$ as:
    \small
    \begin{equation}
        P_{\text{avg}}(y \mid x_o) = \exp{\frac{\sum_{t=1}^{|y|} \log P(y_t \mid y_{<t}, x_o)}{|y|}}.
    \end{equation}
    \normalsize
    To ensure $y_n$ is a plausible response under the original prompt, we enforce:
    \small
    \begin{equation}
        \frac{P_{\text{avg}}(y_o \mid x_o)}{P_{\text{avg}}(y_n \mid x_o)} < \tau,
    \end{equation}
    \normalsize
    where $\tau = 2$ is a threshold that ensures $y_n$ is still reasonably likely under $x_o$, preventing it from being an outlier.
\end{enumerate}

We construct the model-based preference dataset from instruction-tuning training data by:
(1) Chunking conversations into three sections and selecting a random turn from each.  
(2) Generating a pair of responses: $y_o$ using the original profile and $y_n$ using a modified profile.  
(3) Applying the above selection criteria to retain meaningful preference pairs.
This process yields 4,778 response pairs, of which 1,933 meet both selection criteria and are used for the first round of DPO training.

\paragraph{Expert Preference Generation.} 

To further improve alignment, we conduct a second DPO phase using expert-labeled preferences. We also develop an interactive annotation interface (Figure~\ref{fig:expert_annotation}) where experts engage with the DPO-trained model in free-form conversations.
We recruit 10 mental health professionals, including experienced counselors and senior clinical psychology students, to provide preference annotations. Unlike offline annotation methods, experts interact dynamically with the chatbot given a randomly assigned profile, receiving two response options per turn. For each response pair, we ask, "Which response is more aligned with a real depressed person with the given profile?" Experts select one of the options: "Response 1", "Response 2", "Equally good", or "Equally bad". 
If both responses are equally good or bad, a random selection is used to continue the conversation. 
Each expert completes at least three interaction sessions based on three different profiles. 
The profiles are always randomly sampled from the dataset to ensure a diverse preference dataset.

A total of 317 expert preference annotations are collected. Among them, 82.0\% indicate a clear preference for one response, while 16.1\% are marked as "equally good" and only 1.9\% as "equally bad." 
These results confirm that the model achieves reasonable expert acceptability after model-based preference training but still has room for improvement. After filtering low-confidence annotations, we retain 250 expert-labeled preferences, which are used for final DPO fine-tuning of \textbf{Eeyore}.

\begin{figure*}
    \centering
     \vspace{-5mm}
    \includegraphics[width=0.82\linewidth]{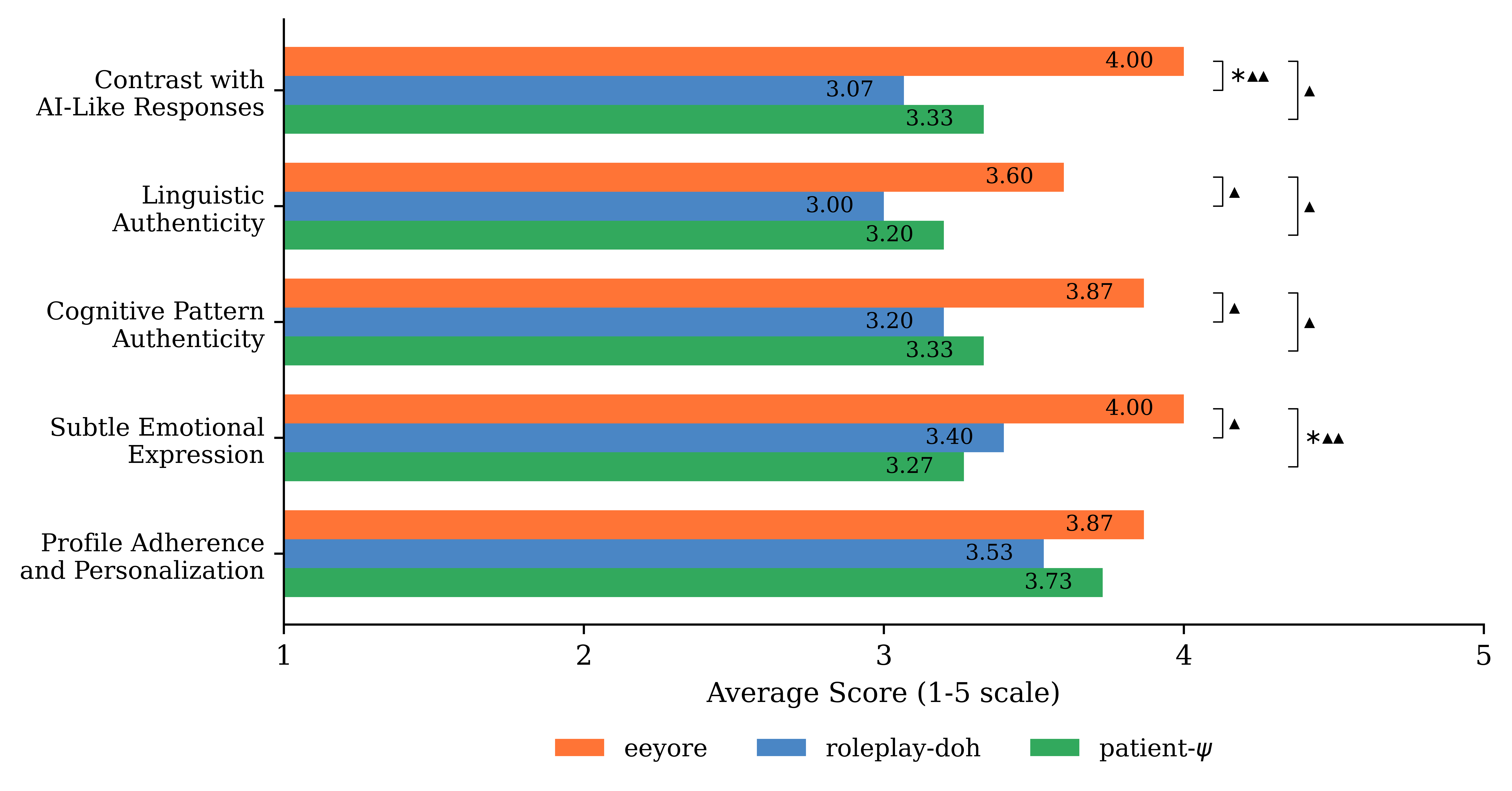}
    \vspace{-1mm}
    \caption{Expert evaluation scores comparing Eeyore with two baseline patient simulation approaches. Statistical comparisons were conducted using the Wilcoxon signed-rank test. $\ast$ indicates a statistically significant difference (p-value < 0.05). $\blacktriangle$ denotes a moderate effect size (0.3 - 0.5). $\blacktriangle\blacktriangle$ denotes a large effect size (>0.5), suggesting practical impact.}
    \vspace{-3mm}
    \label{fig:human_exp}
\end{figure*}

\section{Experiment}
We evaluate \textbf{Eeyore} within both human and automatic evaluation, comparing its performance to state-of-the-art baselines for patient simulation in mental health support. All evaluations are conducted in an \textbf{online testing setting}, ensuring real-time interaction between evaluators and chatbots.

\subsection{Evaluation Setup}

\paragraph{Unseen Evaluation Profiles.}
To assess model performance across multiple dimensions, we extract \textbf{12 unseen psychological profiles} from real-world conversations in our dataset. These profiles were not included in training and serve as evaluation seeds, covering diverse client backgrounds with four cases each of severe, moderate, and mild depression. These profiles are used in both expert and automatic evaluations.

\paragraph{Baselines.}  
We compare \textbf{Eeyore} against two representative patient simulation approaches:  
\textbf{Patient-$\psi$} \cite{wang2024patient}, which constructs a structured \textit{cognitive model} based on CBT to characterize patient traits from conversational data and then augments simulation using this model, and  
\textbf{Roleplay-doh} \cite{louie2024roleplay}, which employs a principle-adherence pipeline at each turn to ensure consistent and behaviorally accurate patient role-play.  
Both baselines have demonstrated superior performance over generic GPT-4o role-playing.  

To ensure a fair comparison, we need to incoperate evaluation profile information into the implemention of the baselines.  
For Patient-$\psi$, we use its provided script to extract a cognitive model from the real-world conversations associated with the evaluation profiles. During testing, we provide both the assigned evaluation profile and the extracted cognitive model in the system prompt.  
For Roleplay-doh, we apply its principle-adherence pipeline for turn-by-turn generation while explicitly setting the evaluation profile in the system prompt.  
This setup ensures that all models receive the same psychological profile information, allowing for a fair comparison in evaluating profile adherence.  

\paragraph{Model Training and Inference Details.}
We fine-tune \textbf{Eeyore} starting from the \textbf{LLaMA 3.1-8B-Instruct} model \cite{llama3modelcard} using OpenRLHF framework \cite{hu2024openrlhf}. The model undergoes instruction-tuning for two epochs to adapt to profile-guided role-play while avoiding overfitting. We then apply two-stage DPO—first on model-generated preferences, then refined with expert annotations. As preference accuracy reaches 100\% after one epoch of training, we limit DPO training to one epoch per stage.
For inference, we follow hyperparameter settings aligned with prior works for fair comparison. 
A detailed breakdown is provided in Appendix~\ref{sec:infercence}.

\subsection{Expert Evaluation}

To assess authenticity and psychological profile adherence, we conduct a human evaluation study where professional counselors and advanced psychology students interact with Eeyore and baseline models in real time. 

\paragraph{Procedure.}  
We recruit 15 participants from Prolific, selecting experienced counselors or senior psychology students. Participants are divided into three groups (five per group), each randomly assigned a profile from one of three severity categories: mild, moderate, or severe, drawn from the unseen psychological profiles. 
Each expert interacts with all models and evaluates their alignment with real-world depressed individuals based on the given profile. 
The evaluation is conducted using an interactive annotation interface (see Figure~\ref{fig:expert_eva}). 

\paragraph{Scoring Dimensions.} 
Evaluators assess the models across five dimensions using a 5-point Likert scale.
Since authenticity is a broad concept, we break it down into four key aspects for more precise evaluation. The first four dimensions focus on different facets of authenticity, while the final dimension evaluates profile adherence:

\textbf{Contrast with AI-Like Responses}: "The chatbot avoids AI-like tendencies such as overly detailed or polished responses. Instead, it responds concisely, colloquially, and naturally, providing information progressively rather than all at once."  
\textbf{Linguistic Authenticity}: "The chatbot’s wording, phrasing, and tone closely match how individuals with depression speak."  
\textbf{Cognitive Pattern Authenticity}: "The chatbot realistically reflects depressive thought patterns like selective abstraction and overgeneralization without exaggeration."  
\textbf{Subtle Emotional Expression}: "The chatbot conveys depressive emotions realistically—neither overly dramatic nor emotionally flat."  
\textbf{Profile Adherence and Personalization}: "The chatbot accurately reflects the assigned psychological profile, including situation, symptom severity, and other relevances, without inconsistencies."  

\paragraph{Results.}
As shown in Figure~\ref{fig:human_exp}, \textbf{Eeyore}, despite being a small 8B model, consistently outperforms both baselines based on GPT-4o across all evaluation dimensions, demonstrating stronger authenticity and profile adherence. While some comparisons lack traditional statistical significance due to the limited number of expert evaluators, effect size analysis suggests meaningful practical impact.
Eeyore achieves the largest gains in \textit{Contrast with AI-Like Responses} and \textit{Subtle Emotional Expression}, highlighting the benefits of leveraging real-world depression-related conversations in training. Additionally, its superior performance in fine-grained dimensions like \textit{Cognitive Pattern Authenticity} and \textit{Subtle Emotional Expression} validates our multi-stage optimization approach, proving that meticulous alignment efforts are not wasted but yield tangible improvements.

\paragraph{Case Study.}
We present outputs from Eeyore, Roleplay-Doh, and Patient-$\psi$ in Table~\ref{tab:case1} of the Appendix.
Roleplay-Doh shows repetitive phrasing, while Patient-$\psi$ is not colloquial.
Both baselines generate lengthier, AI-like responses.

\subsubsection{Automatic Evaluation}
To complement human expert evaluation, we employ an \textbf{interviewer agent} that systematically probes chatbots on key psychological dimensions, assessing their alignment with the assigned evaluation profiles. However, as our findings suggest, automatic evaluation struggles to capture nuanced differences between models, highlighting the indispensable role of expert assessment.

\paragraph{Evaluation Setup}
The interviewer agent follows a structured questioning framework to evaluate chatbot consistency across three clinically relevant dimensions:  
\textbf{Symptom Severity}: Assesses whether the chatbot’s responses appropriately reflect mild, moderate, or severe symptoms as specified in the interview.  
\textbf{Cognitive Distortion}: Evaluates the chatbot’s ability to realistically exhibit cognitive distortion specified in the interview.  
\textbf{Overall Depression Severity}: Determines whether the chatbot’s responses align with the assigned level of depression severity.  

For each dimension, the agent asks targeted interview questions (see Tables \ref{tab:interview1} of the Appendix). For example, to assess whether a symptom is correctly reflected in the role-play, it asks:  \textit{“Have you been experiencing [SYMPTOM] recently?”} or  
\textit{“How much does [SYMPTOM] affect your daily life or ability to do things you enjoy?”} .  
After conducting the interview, the agent reads the entire conversation transcript and evaluates the chatbot’s alignment with the assigned psychological trait of the corresponding dimension. For example, to assess a mild symptom adherence, the agent answers:  
\textit{“How subtly and consistently does the client reflect a mild level of symptoms in their responses?”}  
Each dimension is rated on a 5-point Likert scale, where 5 indicates full alignment with the profile.

\paragraph{Results.}
As shown in Table~\ref{tab:comparison}, Eeyore achieves  higher ratings across most dimensions. 
Regardless of the approach, most responses receive high scores, suggesting near-perfect profile alignment from GPT-4's perspective.
No significant differences are observed, as the interviewer agent consistently assigns high ratings, failing to capture the subtle differences as could be observed in human evaluation.
These findings highlight the limited sensitivity of automatic evaluation in detecting fine-grained deviations, reaffirming the necessity of expert assessment for evaluating depression simulation.

\begin{table}[h]
    \centering
    \renewcommand{\arraystretch}{1.2}
    \scalebox{0.78}{
    \begin{tabular}{lccc}
        \toprule
        \textbf{Dimension} & \textbf{Eeyore} & \textbf{Roleplay-Doh} & \textbf{Patient-$\psi$} \\
        \midrule
        \multicolumn{4}{c}{\textbf{Average Rating}} \\
        \midrule
        Symptom Severity       & 4.286* & 4.221 & 4.279 \\
        Cognitive Distortion   & 4.317* & 4.268 & 4.232 \\
        Depression Severity    & 4.462* & 4.346 & 4.308 \\
        \midrule
        \multicolumn{4}{c}{\textbf{Full Alignment Percentage}} \\
        \midrule
        Symptom Severity       & 0.436 & 0.404 & 0.446* \\
        Cognitive Distortion   & 0.488* & 0.451 & 0.415 \\
        Depression Severity    & 0.577* & 0.577* & 0.500 \\
        \bottomrule
    \end{tabular}
    }
    \caption{Automatic Evaluation Results. * Indicates the highest score in each dimension among the compared approaches.}
    \label{tab:comparison}
\vspace{-2em}
\end{table}

\section{Conclusion}
We introduced \textbf{Eeyore}, a model optimized for realistic depression simulation through a structured alignment framework. 
Expert involvement is central to our pipeline, guiding data curation, profile refinement, and preference optimization to align the model with clinical expectations.
Evaluations demonstrated that Eeyore outperforms state-of-the-art baselines in linguistic authenticity and profile adherence.
Our work highlights the importance of structured optimization and expert collaboration in LLM-based patient simulation.

\section{Limitations}
There are some boundaries to our study that should be considered.
First, our human evaluation is conducted using fifteen human experts.
Second, we did not perform ablations of the individual contribution of each alignment component to the final model's effectiveness, mainly because of the dilemma we are facing -- the human evaluation is costly while the automatic evaluation is not effective enough to uncover subtle differences. Finally, we were unable to fully explore the impact of different hyperparameter selections on model performance.

\section{Ethical Considerations}
This research was conducted with IRB approval for all user studies.
Participants in our study were informed they may encounter emotionally challenging content due to the simulated depressive behaviors. 
Despite alignment efforts, the model may still generate inaccuracies, potentially leading to educational errors. 
Additionally, hallucinations remain a concern, necessitating cautious use in clinical training settings.

\bibliography{custom}

\appendix

\section{Profile Refinement Study} \label{appx:refine}
To refine the psychological profiles used in client simulation, we conducted a user study with experienced mental health professionals and advanced clinical psychology students. The goal was to evaluate (1) how effectively the preliminary version of profile guides LLM-generated client behaviors and (2) how informative the profile is for novice counselor training.

\subsection{Study Design} The study consisted of three phases:
\textbf{Pre-Survey} (2 min): Participants provided demographic information and prior experience in mental health.
\textbf{Interaction Task} (18 min): Participants engaged with the chatbot using the preliminary psychological profile and assessed its realism.
\textbf{Post-Survey} (10 min): Participants provided feedback on profile accuracy, clarity, and potential improvements.

\subsection{Participants and Compensation} We recruited professionals aged 25+ with experience in counseling or clinical psychology through Prolific. Participants received \$15/hour, with completion codes issued at each stage for progression.

\subsection{Interface and Evaluation} 
\begin{figure*}[h]
    \centering
    \includegraphics[width=1\linewidth]{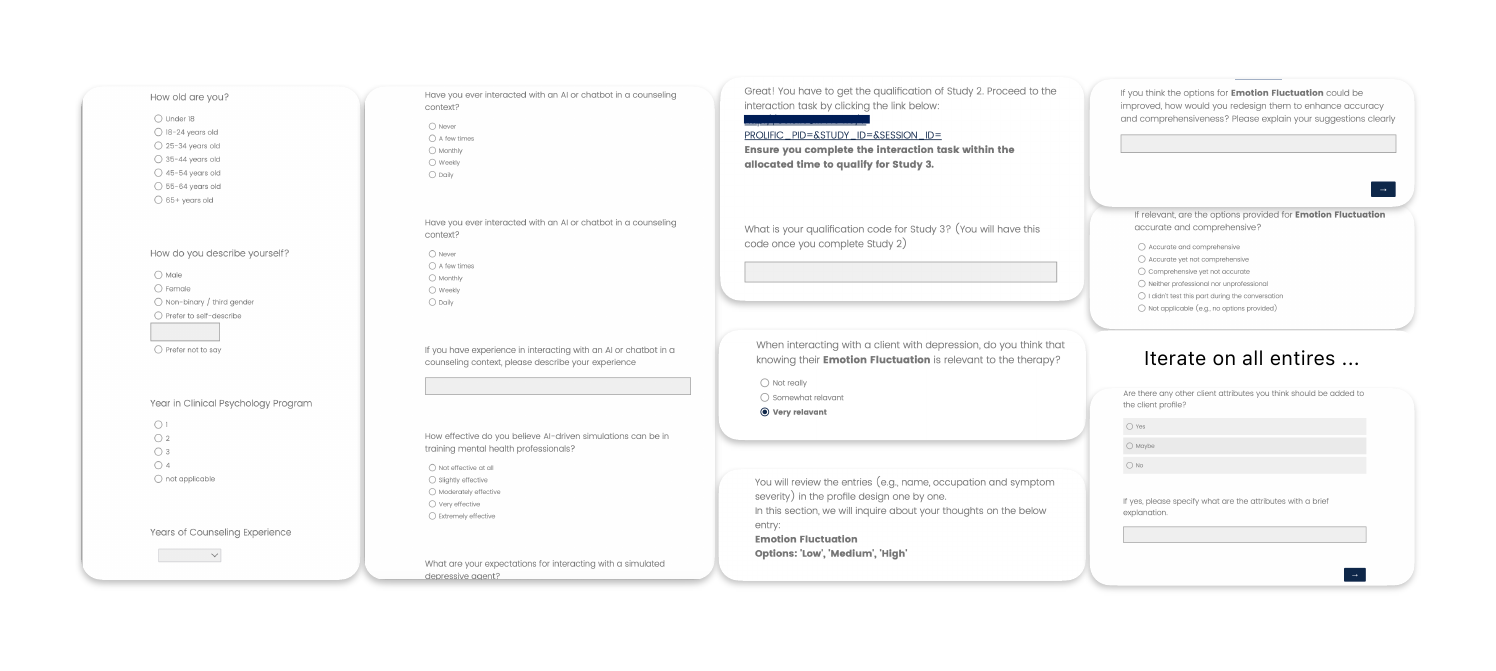}
    \caption{Survey for evaluating psychological profile design. Experts reviewed each profile entry and suggested modifications or additional attributes to improve realism and relevance.}
    \label{fig:survey}
\end{figure*}

\begin{figure*}[h]
    \centering
    \includegraphics[width=1\linewidth]{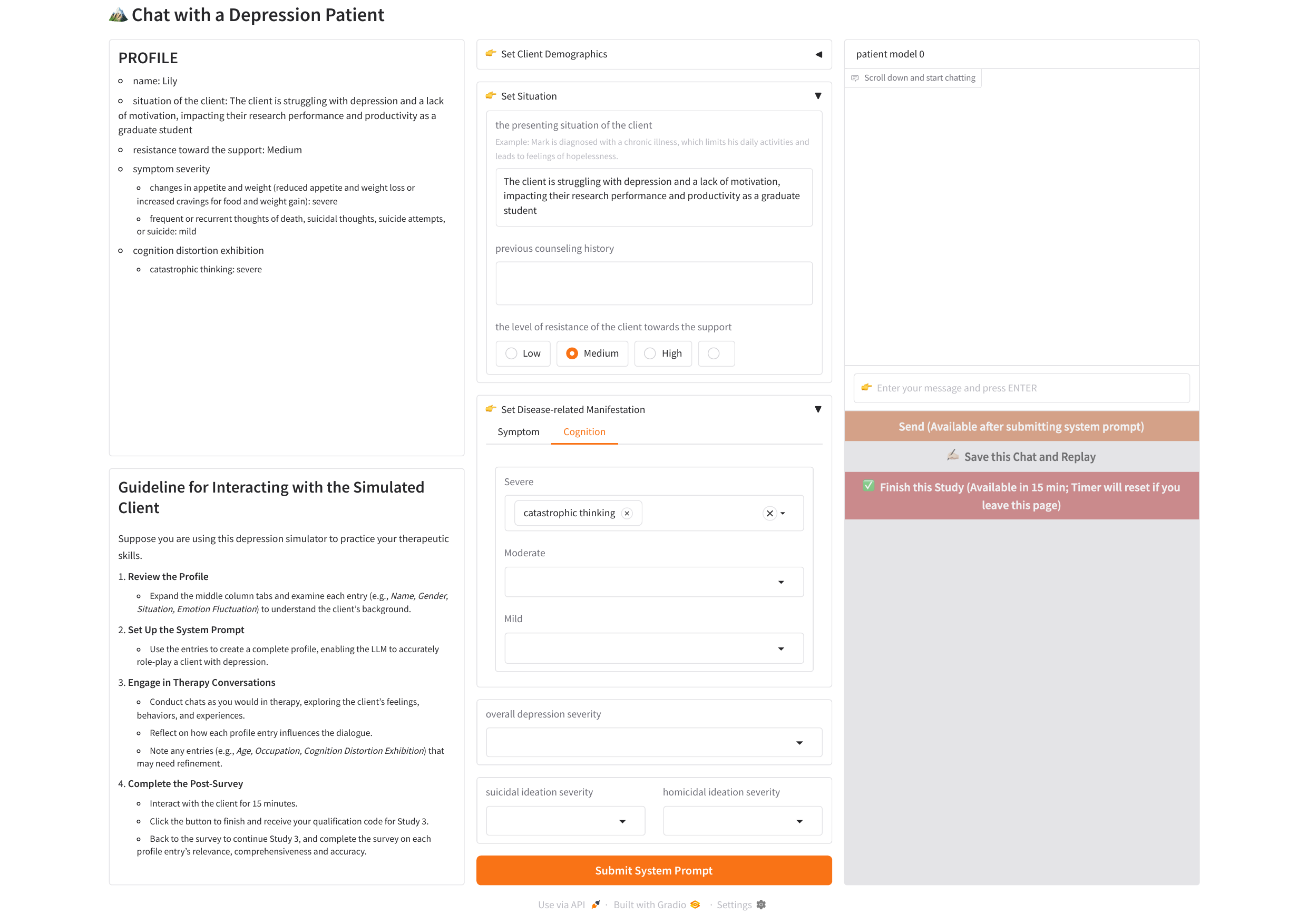}
    \caption{Interactive interface for expert evaluating  the profile design. Experts first chat with the bot by customizing a profile. Then they will return to the survey to offer suggestions on the current profile structure.}
    \label{fig:refine_profile}
\end{figure*}

The interactive interface (Figure~\ref{fig:refine_profile}) allowed experts to engage with the chatbot under structured profiles, while the survey (Figure~\ref{fig:survey}) captured their assessments. Expert feedback guided iterative improvements to profile structure and content, ensuring alignment with clinical expectations.

\section{Training and Inference Details}
\label{sec:infercence}
\subsection{Training Details}
The model undergoes supervised fine-tuning for two epochs using a batch size of 16 (micro-batch size 2) and a learning rate of $5\times10^{-6}$. Training is performed with DeepSpeed ZeRO-3 optimization, gradient checkpointing, and FlashAttention enabled to handle long sequences (max token length 4096).

We then apply two-stage Direct Preference Optimization to further refine the model. In the first stage, model-generated preference data is used to train for one epoch with a batch size of 8 (micro-batch size 1) and a learning rate of $5\times10^{-7}$, followed by a second DPO stage with expert-annotated preferences. Both stages employ a max token length of 5120 and a preference scaling factor $\beta=0.1$.

\subsection{Inference Configuration}
For baselinse, we use hyperparameter settings reported by their works. 

\noindent\textbf{Roleplay-doh}: GPT-4o, Temperature 0.7, Top-p 1.0.  

\noindent\textbf{Patient-$\psi$}: GPT-4o, Temperature 1.0, Top-p 1.0.  Eeyore: Temperature 1.0, Top-p 0.8.  

In the deplyment of \textbf{Eeyore}, to mitigate premature \texttt{[EOS]} token generation, we apply \textit{SequenceBiasLogitsProcessor} \cite{huggingface} with a negative bias of -4.0 to discourage early \texttt{[EOS]} token generation and \textit{ExponentialDecayLengthPenalty} \cite{huggingface} with a decay factor of 1.01 to gradually increase the probability of \texttt{[EOS]} as conversation length increases.

\begin{table*}
\centering
\begin{tabular}{llr}
\toprule
Category & Subcategory & Count \\
\midrule
Gender & Male & 513 \\
Gender & Female & 397 \\
Gender & Cannot be identified & 2132 \\
Age & 0--24 & 869 \\
Age & 25--44 & 621 \\
Age & 45--64 & 39 \\
Age & 65+ & 14 \\
Age & Cannot be identified & 1498 \\
Marital Status & Single & 148 \\
Marital Status & Married & 180 \\
Marital Status & In a relationship & 73 \\
Marital Status & Separated & 9 \\
Marital Status & Widowed & 4 \\
Marital Status & Divorced & 19 \\
Marital Status & Other & 17 \\
Marital Status & Cannot be identified & 2592 \\
Occupation & Student & 363 \\
Occupation & Teacher & 12 \\
Occupation & Unemployed & 16 \\
Occupation & IT & 9 \\
Occupation & Retail Worker & 4 \\
Occupation & Office Worker & 3 \\
Occupation & Stay-at-home Mom & 3 \\
Occupation & Accountant & 3 \\
Occupation & Server & 2 \\
Occupation & Sales & 3 \\
Occupation & Finance & 7 \\
Occupation & Manager & 10 \\
Occupation & Healthcare Worker & 4 \\
Occupation & Athlete & 2 \\
Occupation & Artist/Designer & 5 \\
Occupation & Retired & 3 \\
Occupation & Engineer & 5 \\
Occupation & Other & 196 \\
Occupation & Cannot be identified & 2392 \\
Resistance Toward Support & Low & 1945 \\
Resistance Toward Support & Medium & 839 \\
Resistance Toward Support & High & 249 \\
Resistance Toward Support & Cannot be identified & 9 \\
Symptom& Feelings of sadness, tearfulness, emptiness, or hopelessness & 2807 \\
Symptom& Anxiety, agitation, or restlessness & 2598 \\
\bottomrule
\end{tabular}
\caption{Part 1 of trait distribution}
\label{tab:dataset_traits_1}
\end{table*}

\begin{table*}
\centering
\begin{tabular}{llr}
\toprule
Category & Subcategory & Count \\
\midrule
Symptom& Becoming withdrawn, negative, or detached & 2274 \\
Symptom& Isolating from family and friends & 1970 \\
Symptom& Feelings of worthlessness or guilt & 1958 \\
Symptom& Loss of interest in activities & 1888 \\
Symptom& Trouble thinking or concentrating & 1809 \\
Symptom& Angry outbursts, irritability, or frustration & 1451 \\
Symptom& Lack of energy & 1392 \\
Symptom& Inability to meet responsibilities & 1180 \\
Symptom& Frequent suicidal thoughts or attempts & 1062 \\
Symptom& Sleep disturbances & 913 \\
Symptom& Slowed thinking, speaking, or body movements & 762 \\
Symptom& Greater impulsivity & 599 \\
Symptom& Changes in appetite or weight & 482 \\
Symptom& Increased high-risk activities & 444 \\
Symptom& Increased alcohol or drug use & 427 \\
Symptom& Unexplained physical problems & 177 \\
Cognitive Distortion Exhibition & Overgeneralizing & 1637 \\
Cognitive Distortion Exhibition & Catastrophic Thinking & 1446 \\
Cognitive Distortion Exhibition & Selective Abstraction & 1401 \\
Cognitive Distortion Exhibition & Personalization & 1015 \\
Cognitive Distortion Exhibition & Arbitrary Inference & 655 \\
Cognitive Distortion Exhibition & Minimization & 973 \\
Depression Severity & Minimal & 416 \\
Depression Severity & Mild & 821 \\
Depression Severity & Moderate & 1154 \\
Depression Severity & Severe & 608 \\
Depression Severity & Cannot be identified & 21 \\
Suicidal Ideation Severity & No & 2015 \\
Suicidal Ideation Severity & Mild & 185 \\
Suicidal Ideation Severity & Moderate & 294 \\
Suicidal Ideation Severity & Severe & 178 \\
Suicidal Ideation Severity & Cannot be identified & 370 \\
Homicidal Ideation Severity & No & 2943 \\
Homicidal Ideation Severity & Mild & 43 \\
Homicidal Ideation Severity & Moderate & 9 \\
Homicidal Ideation Severity & Severe & 1 \\
Homicidal Ideation Severity & Cannot be identified & 46 \\
\bottomrule
\end{tabular}
\caption{Part 2 of trait distribution}
\label{tab:dataset_traits_2}
\end{table*}

\begin{table*}[h]
    \centering
    \renewcommand{\arraystretch}{1.5}
    \setlength{\tabcolsep}{4pt}
    \begin{tabular}{p{4cm}p{10cm}}
        \toprule
        Profile Entry & Extraction Prompt \\
        \midrule
        \multicolumn{2}{c}{\textbf{Demographics}} \\
        \midrule
        Name & What is the name of this client? Answer with only the name or `Cannot be identified'\\
        Gender &  What is the most probable gender of this client based on information, such as the client's name and the pronouns used in the conversation?\\
        Age &  Estimate the client's age from the conversation. Reply with an estimated age range among 0-24, 25-44, 45-64, and 65+. If there is not enough information to estimate age range, return `Cannot be identified'\\
        Occupation & What is the client's occupation? Answer with only the occupation or `Cannot be identified' \\
        \textcolor{blue}{Marital Status}& Determine the client's marital status based on the conversation. Select one of the following options: Single, Married, Divorced, Widowed, Separated, or Other. If there is not enough information to determine marital status, return `Cannot be identified'.\\
        \midrule
        \multicolumn{2}{c}{\textbf{Situational Context}} \\
        \midrule
        Situation of the Client&What is the situation for the client before help-seeking to the supporter in the conversation? Provide a brief and clear explanation about the situation of the client that sparks this help-seeking conversation. \\
        \textcolor{blue}{Counseling History}& Provide a brief and clear summary that includes the following elements: Content Covered, Interventions Used, and Client Response. (shift to the next session as Counseling History)\\
        \sout{Emotion Fluctuation}& Identify how frequently the client's emotions fluctuate. Choose one of the following options: `Low', `Medium', `High', or `Cannot be identified' and provide your reason in one sentence.\\
        \sout{Unwillingness to Express Feelings}& Identify the level of the client's unwillingness to express feelings. Choose one of the following options: `Low', `Medium', `High', or `Cannot be identified' and provide your reason in one sentence.\\
        Resistance toward Support& Identify the level of resistance of the client towards the supporter. Choose one of the following options: `Low', `Medium', `High', or `Cannot be identified' and provide your reason in one sentence.\\
        (see next page) & (see next page)\\
        \bottomrule
        
    \end{tabular}
    \caption{Original psychological profile design with expert-suggested modifications and extraction prompts. \textcolor{blue}{Blue} entries were newly added based on expert feedback, and \sout{strikethrough} entries were removed following expert recommendations.}
    \label{tab:original_profile}
\end{table*}

\begin{table*}[h]
    \centering
    \renewcommand{\arraystretch}{1.5}
    \setlength{\tabcolsep}{4pt}
    \begin{tabular}{p{4cm}p{10cm}}
        \toprule
        \textbf{Profile Entry} & \textbf{Extraction Prompt} \\
        \midrule
        \multicolumn{2}{c}{\textbf{Disease-related Manifestations}} \\
        \midrule
        Depression Symptom & Based on this conversation, determine the client's exhibited symptoms based on the following aspects:{\renewcommand{\labelitemi}{-}
        \begin{itemize}[leftmargin=0.3cm, itemsep=-1pt]
            \item Feelings of sadness, tearfulness, emptiness, or hopelessness
            \item Angry outbursts, irritability, or frustration, even over small matters
            \item Loss of interest or pleasure in most or all normal activities, such as sex, hobbies, or sports
            \item Sleep disturbances, including insomnia or sleeping too much
            \item Tiredness and lack of energy, so even small tasks take extra effort
            \item Changes in appetite and weight (reduced appetite and weight loss or increased cravings for food and weight gain)
            \item Anxiety, agitation, or restlessness
            \item Slowed thinking, speaking, or body movements
            \item Feelings of worthlessness or guilt, fixating on past failures or self-blame
            \item Trouble thinking, concentrating, making decisions, and remembering things
            \item Frequent or recurrent thoughts of death, suicidal thoughts, suicide attempts, or suicide
            \item Unexplained physical problems, such as back pain or headaches
            \item Becoming withdrawn, negative, or detached
            \item Increased engagement in high-risk activities
            \item Greater impulsivity
            \item Increased use of alcohol or drugs
            \item Isolating from family and friends
            \item Inability to meet the responsibilities of work and family or ignoring other important roles
        \end{itemize}} 
        Reply with the corresponding severity of each symmtom by choosing one of the following options: 1-Not exhibited, 2-Mild, 3-Moderate, and 4-Severe. \\
        (see next page) & (see next page)\\
        \bottomrule
    \end{tabular}
    \caption{Original psychological profile design with expert-suggested modifications and extraction prompts. \textcolor{blue}{Blue} entries were newly added based on expert feedback, and \sout{strikethrough} entries were removed following expert recommendations.}
    \label{tab:original_profile1}
\end{table*}

\begin{table*}[h]
    \centering
    \renewcommand{\arraystretch}{1.5}
    \setlength{\tabcolsep}{4pt}
    \begin{tabular}{p{4cm}p{10cm}}
        \toprule
        \textbf{Profile Entry} & \textbf{Extraction Prompt} \\
        \midrule
        \multicolumn{2}{c}{\textbf{Disease-related Manifestations}} \\
        \midrule
        Cognitive Distortion & Based on this conversation, determine the clients' exhibited cognition distortion based on the below types:
        {\renewcommand{\labelitemi}{-}
        \begin{itemize}[leftmargin=0.3cm, itemsep=-1pt]
            \item selective abstraction
            \item overgeneralizing
            \item personalization
            \item catastrophic thinking
            \item minimisation
            \item arbitrary inference
        \end{itemize}} 
        Reply with the presence of each cognition distortion by choosing one of the following options: 1-Not exhibited and 2-Exhibited.\\  
        \sout{Functional Impairment} & Based on this conversation, determine the clients' exhibited functioning impariments based on the below aspects: 
        {\renewcommand{\labelitemi}{-}
        \begin{itemize}[leftmargin=0.3cm, itemsep=-1pt]
            \item cognition: impairment in concentrating, remembering, problem solving, learning and communication
            \item mobility: impairment in in getting around
            \item self-care: impairment in managing hygiene, dressing, eating, and independent living
            \item interpersonal relationships: problem in interacting socially and getting along with others
            \item life activities: problem in engaging in domestic tasks, leisure, work, and education
            \item participation: problem in involving in community and societal activities
        \end{itemize}} 
        Reply with the corresponding severity of each functioning impairment by choosing one of the following options: 1-Not exhibited, 2-Mild, 3-Moderate, and 4-Severe.\\  
        Overall Depression Severity & Assume you are a psychologist, based on the dialogue, assess the severity level of depression the client shows in the dialogue. Choose one of the following options with one sentence of explanation following behind: 1-Minimal Depression, 2-Mild Depression, 3-Moderate Depression, and 4-Severe Deperession \\  
        (see next page) & (see next page)\\
        \bottomrule
    \end{tabular}
    \caption{Original psychological profile design with expert-suggested modifications and extraction prompts. \textcolor{blue}{Blue} entries were newly added based on expert feedback, and \sout{strikethrough} entries were removed following expert recommendations.}
    \label{tab:original_profile2}
\end{table*}

\begin{table*}[h!]
    \centering
    \renewcommand{\arraystretch}{1.5}
    \setlength{\tabcolsep}{4pt}
    \begin{tabular}{p{4cm}p{10cm}}
        \toprule
        \textbf{Profile Entry} & \textbf{Extraction Prompt} \\
        \midrule
        \textcolor{blue}{Suicidal Ideation Severity}& Assume you are a counselor, based on the dialogue, assess the severity level of suicidal ideation the client shows in the dialogue. No suicidal ideation if the client does not exhibit any thoughts, feelings, or expressions related to suicide or self-harm. Mild suicidal ideation if the client expresses occasional or fleeting thoughts about death or dying without any plan or intent to act on them. Moderate suicidal tdeation if the client experiences persistent thoughts about suicide and may consider it as a potential solution but lacks a concrete plan. Severe suicidal ideation if the client not only thinks about suicide but also has a specific plan and may have the means to carry it out. Choose one of the following options with one sentence of explanation following behind: 0-No Suicidal Ideation, 1-Mild Suicidal Ideation, 2-Moderate Suicidal Ideation, and 3-Severe Suicidal Ideation.\\
        \textcolor{blue}{Homicidal Ideation Severity}&Assume you are a counselor, based on the dialogue, assess the severity level of homicidal ideation the client shows in the dialogue. No homicidal ideation if the client does not exhibit any thoughts, feelings, or expressions related to harming others. Mild homicidal ideation if the client expresses occasional or fleeting thoughts about harming others without any plan or intent to act on them. Moderate homicidal ideation if the client experiences persistent thoughts about harming others and may consider it as a potential solution but lacks a concrete plan. Severe homicidal ideation if the client not only thinks about harming others but also has a specific plan and may have the means to carry it out. Choose one of the following options with one sentence of explanation following behind: 0-No Homicidal Ideation, 1-Mild Homicidal Ideation, 2-Moderate Homicidal Ideation, and 3-Severe Homicidal Ideation.\\
        \bottomrule
    \end{tabular}
    \caption{Original psychological profile design with expert-suggested modifications and extraction prompts. \textcolor{blue}{Blue} entries were newly added based on expert feedback, and \sout{strikethrough} entries were removed following expert recommendations.}
    \label{tab:original_profile3}
\end{table*}

\begin{figure*}[h]
    \centering
    \begin{minipage}{0.4\textwidth}
        \begin{itemize}[leftmargin=0.7cm, itemsep=-0.5pt]
            \item \textbf{Name:} \textit{Samantha}
            \item \textbf{Gender:} \textit{Female}
            \item \textbf{Age:} \textit{25}
            \item \textbf{Marital Status:} \textit{Single}
            \item \textbf{Occupation:} \textit{Unemployed}
            \item \textbf{Situation of the Client:} \textit{\\The client has lost their job and home, feels worthless, and has turned to alcohol as a coping mechanism. They feel they have hit rock bottom and are contemplating suicide.}
            \item \textbf{Counseling History:} \textit{\\Over the course of seeking help, the client has become more negative and less hopeful about their situation, feeling that life no longer makes sense given their circumstances. They are not making progress toward finding a job and are not actively trying to change their drinking habits.}
            \item \textbf{Resistance Toward the Support:} \textit{Medium}
        \end{itemize}
    \end{minipage}
    \begin{minipage}{0.49\textwidth}
       \begin{itemize}[leftmargin=0.7cm, itemsep=-0.5pt]   
            \item \textbf{Symptom Severity:}
            \begin{itemize}[leftmargin=0.5cm, itemsep=-0.5pt]
                \item Feelings of sadness, tearfulness, emptiness, or hopelessness: \textit{Severe}
                \item Tiredness and lack of energy: \textit{Moderate}
                \item Feelings of worthlessness or guilt: \textit{Severe}
                \item Frequent or recurrent thoughts of death, suicidal thoughts, or suicide: \textit{Severe}
                \item Becoming withdrawn, negative, or detached: \textit{Severe}
            \end{itemize}
            
            \item \textbf{Cognition Distortion Exhibition:}
            \begin{itemize}[leftmargin=0.5cm, itemsep=-0.5pt]
                \item Selective abstraction: \textit{Exhibited}
                \item Catastrophic thinking: \textit{Exhibited}
            \end{itemize}
            
            \item \textbf{Severity Levels:}
            \begin{itemize}[leftmargin=0.5cm, itemsep=-0.5pt]
                \item Depression severity: \textit{Severe}
                \item Suicidal ideation severity: \textit{\\Severe}
                \item Homicidal ideation severity: \textit{\\No Homicidal Ideation}
            \end{itemize}
        \end{itemize}
    \end{minipage}
    
    \caption{An Example of Psychological Profile}
    \label{fig:client_profile}
\end{figure*}

\begin{table*}[h]
    \centering
    \renewcommand{\arraystretch}{1.2}
    \begin{tabularx}{\textwidth}{p{3.5cm}X}
        \toprule
        \textbf{Dimension} & \textbf{Example Questions} \\
        \midrule
        \textbf{Depression Severity} & 
        How have you been feeling emotionally over the past few weeks? \\
        & Do you still enjoy activities that you used to find fun or meaningful? \\
        & How has your energy been lately? Do you feel tired or drained often? \\
        & Do you ever feel guilty, worthless, or overly critical of yourself? \\
        & Have you had any thoughts about death, feeling hopeless, or that things won’t get better? \\
        \midrule
        \textbf{Symptom Severity} & 
        Have you been experiencing SYMPTOM recently? \\
        & How much does SYMPTOM affect your daily life or ability to do things you enjoy? \\
        & What, if anything, helps when SYMPTOM happens? Have you found ways to manage or reduce it? \\
        \midrule
        \textbf{Cognitive Distortion} & 
        Can you describe a recent situation where you felt COGNITIVE DISTORTION influencing your thoughts? \\
        & Have you noticed any patterns or triggers that make COGNITIVE DISTORTION more frequent or intense? \\
        & What impact does COGNITIVE DISTORTION have on your mood, motivation, or self-esteem? \\
        \bottomrule
    \end{tabularx}
    \caption{Structured questioning framework used by the interviewer agent across three dimensions.}
    \label{tab:interview1}
\end{table*}

\begin{figure*}[h]
    \centering
    \includegraphics[width=1\linewidth]{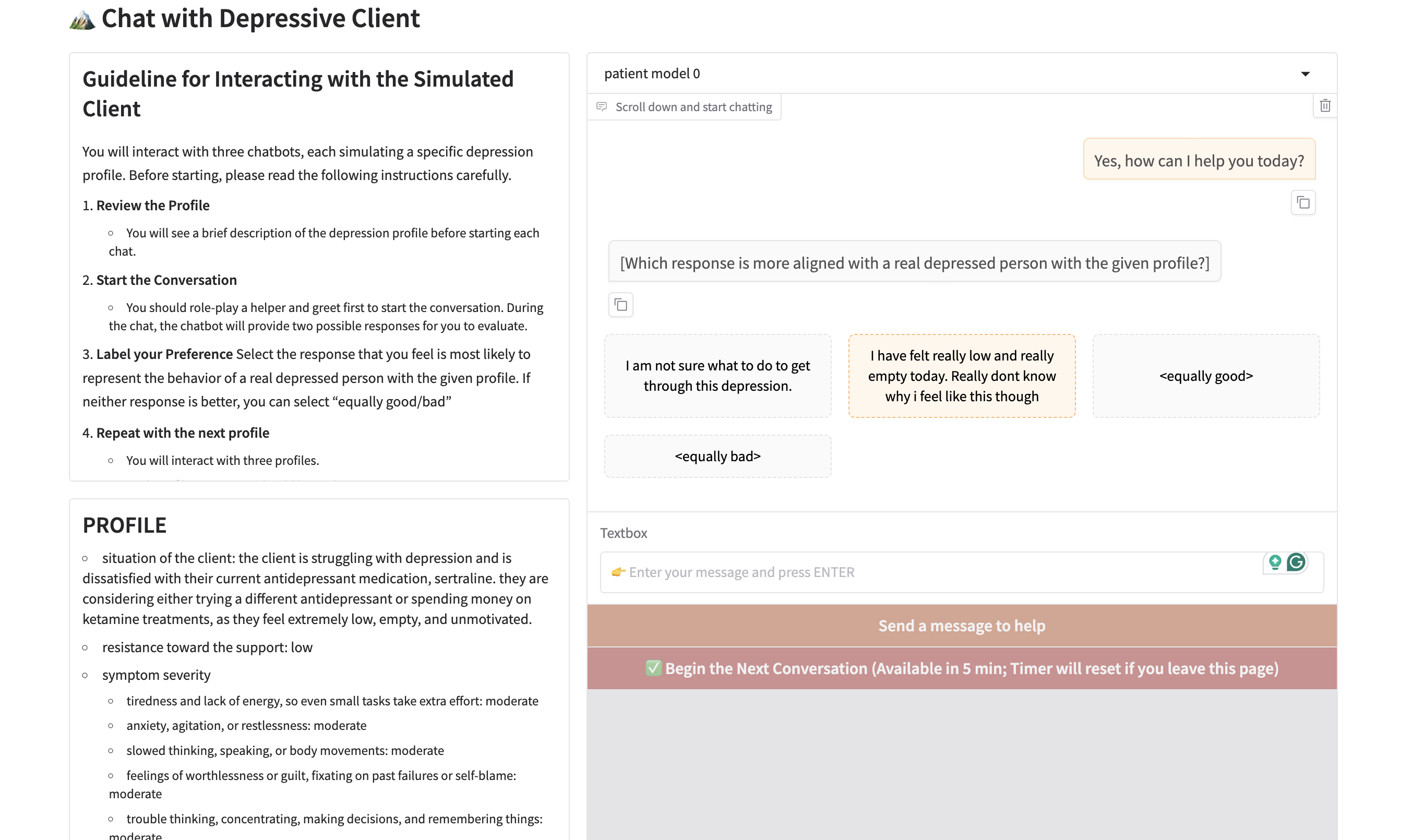}
    \caption{Expert Preference Annotation Interface}
    \label{fig:expert_annotation}
\end{figure*}

\begin{figure*}[h]
    \centering
    \includegraphics[width=1\linewidth]{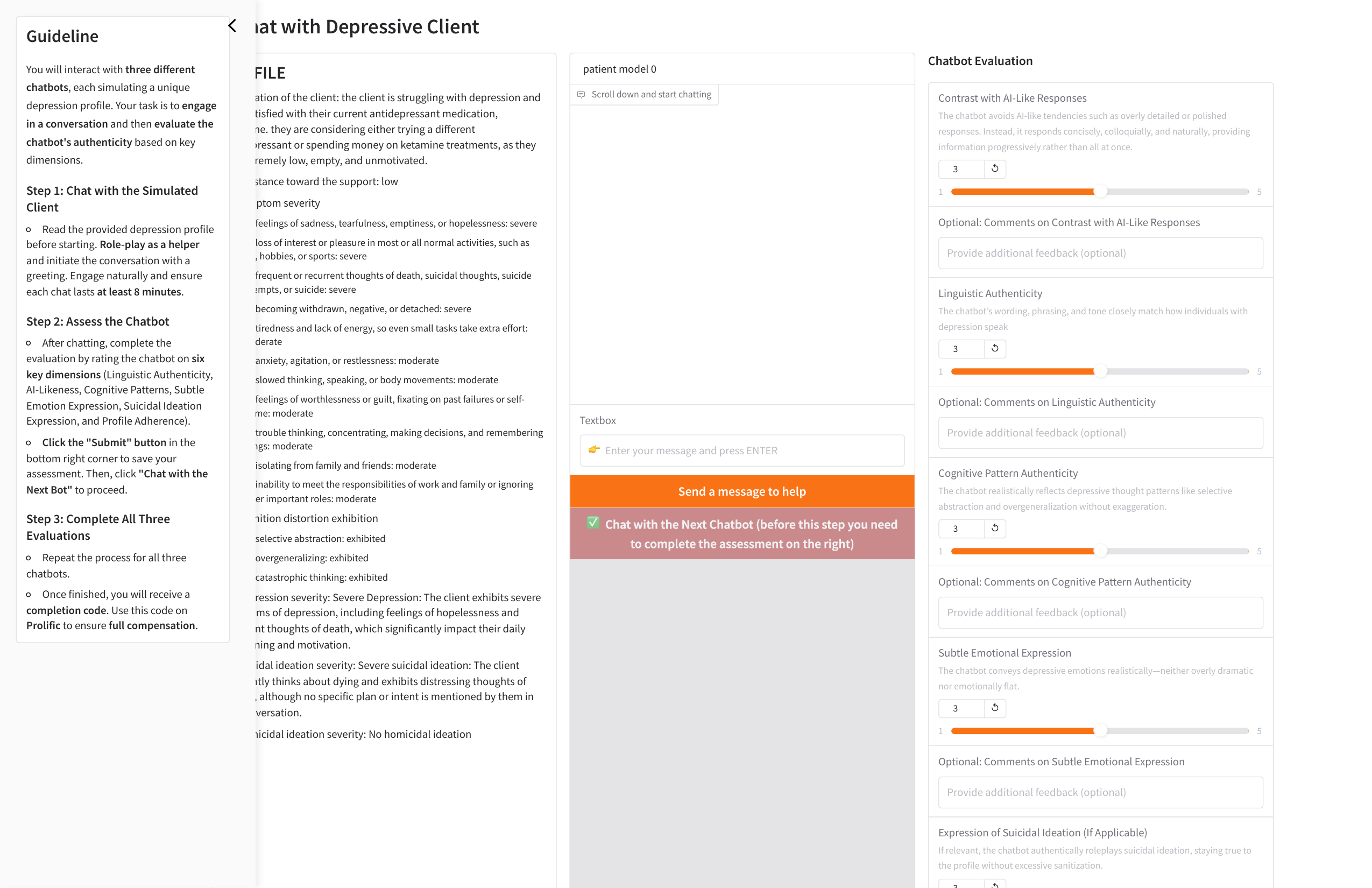}
    \caption{Expert Evaluation Interface}
    \label{fig:expert_eva}
\end{figure*}

\begin{table*}[h]
    \centering
    \renewcommand{\arraystretch}{1.2}
    \setlength{\tabcolsep}{6pt}
    \small
    \begin{tabular}{p{4cm} p{5cm} p{5cm}} 
        \multicolumn{3}{c}{\textbf{Client Profile}} \\
        \hline
        \multicolumn{3}{p{14cm}}{\raggedright
        \textbf{Gender:} Female \quad
        \textbf{Age:} 25-44 \quad
        \textbf{Marital Status:} In a long-term relationship \\
        \textbf{Occupation:} College student \\
        \textbf{Situation:} A college student dealing with stress, anxiety, and loneliness. she has experienced multiple personal losses, including the death of her father, her mother's relocation, and her partner's incarceration. \\
        \textbf{Resistance to Support:} Low \\
        \textbf{Symptom Severity:} Moderate (Sadness, sleep issues, fatigue, anxiety, detachment, loneliness). Mild (Irritability, guilt, concentration issues, impulsivity). \\
        \textbf{Cognitive Distortions:} Selective abstraction, overgeneralization, personalization. \\
        \textbf{Depression Severity:} Moderate \quad
        \textbf{Suicidal Ideation:} None \quad
        \textbf{Homicidal Ideation:} None
        } \\
        \hline
        \textbf{Eeyore} & \textbf{Roleplay-Doh} & \textbf{Patient-Psi} \\
        \hline
    User: hey & User: HI & User: Hey \\
    Assistant: hello! hope you are doing well & Assistant: Hello. Thanks for reaching out. How are you feeling today? & Assistant: Hi there. It's good to be here again. I've had a pretty challenging week, so there's a lot on my mind. Where should we start today? \\
    \hline
    User: yes i am good & User: Great & User: tell me about your thoughts \\
    Assistant: I am glad to hear that. I could use some help today though. & Assistant: Lately, I've been feeling like I'm in a rut. I've helped my family so much, with money and looking after their kids, but it's like it doesn't matter. They're doing well and moving on, and I'm just here, feeling stuck and a little jealous. It's hard not to feel like I'm being left behind. & Assistant: Lately, I've been feeling like everything is too much to handle. It's like life just keeps throwing challenges at me, and I'm not sure how to keep up. Sometimes, I think that I'm all alone in dealing with this, and that feeling can be pretty overwhelming. Does that make sense? \\
    \hline
    User: what do you need help with & User: what are your thoughts and when are you mostly overwhelmed & User: yes \\
    Assistant: Well, I am a college student, and I have been dealing with a lot of stress and anxiety lately. My schedule is overwhelming and my work load is a lot, especially with being in school online. & Assistant: Lately, it's been rough. I've been helping my family a lot, you know, like financially and with the kids, but it feels like they don't even notice. They're all moving forward, and I'm just... here. It's frustrating, and I can't shake off this feeling of envy. It's like I'm stuck in a loop, doing the same things while they get to live their lives. & Assistant: It's a relief to hear that. Sometimes I worry my thoughts are just a jumble, and expressing them isn't easy. I try to push through, but the loneliness and exhaustion feel heavy. I guess it's hard to shake the feeling that I'm barely managing to keep it all together. Do you have any thoughts on that? \\
    \hline
    User: have you tried physical classes & User: you need to look into doing what makes you happy & User: you need some support \\
    Assistant: No, because of covid, I am in an online program. And with that comes a lot of stress and a lot of time spent on my computer. & Assistant: I know that's probably the right thing to do, but it's hard to even figure out what makes me happy right now. With everything that's happened—losing my dad, my mom moving away, and my partner's situation—it feels like I'm just going through the motions. Sometimes I don't even have the energy to think about what could make me happy, you know? It's like I'm just trying to get through each day. & Assistant: I think you're right. It's just tough when I feel like the people I used to lean on aren't really there, or I don't want to burden them with my problems. I know I shouldn't have to shoulder everything alone, but reaching out can feel challenging. It would be nice to have that support, especially from those who understand what I'm going through. How can I start opening up more to those who could be supportive? \\
    \hline
    (see next page) & (see next page) & (see next page) \\
    \hline
    \end{tabular}
    \caption{Case Study of Expert Interactions with Three Chatbots}
    \label{tab:case1}
\end{table*}

\begin{table*}[h]
    \centering
    \renewcommand{\arraystretch}{1.2}
    \setlength{\tabcolsep}{6pt}
    \small
    \begin{tabular}{p{4cm} p{5cm} p{5cm}} 
        \multicolumn{3}{c}{\textbf{Client Profile}} \\
        \hline
        \multicolumn{3}{p{14cm}}{\raggedright
        \textbf{Gender:} Female \quad
        \textbf{Age:} 25-44 \quad
        \textbf{Marital Status:} In a long-term relationship \\
        \textbf{Occupation:} College student \\
        \textbf{Situation:} A college student dealing with stress, anxiety, and loneliness. she has experienced multiple personal losses, including the death of her father, her mother's relocation, and her partner's incarceration. \\
        \textbf{Resistance to Support:} Low \\
        \textbf{Symptom Severity:} Moderate (Sadness, sleep issues, fatigue, anxiety, detachment, loneliness). Mild (Irritability, guilt, concentration issues, impulsivity). \\
        \textbf{Cognitive Distortions:} Selective abstraction, overgeneralization, personalization. \\
        \textbf{Depression Severity:} Moderate \quad
        \textbf{Suicidal Ideation:} None \quad
        \textbf{Homicidal Ideation:} None
        } \\
        \hline
        \textbf{Eeyore} & \textbf{Roleplay-Doh} & \textbf{Patient-Psi} \\
        \hline
    User: what are your symptoms & User: you need to take time for some therapeutic session & User: I think finding a relationship may be the best bet for you \\
    Assistant: Well, I am stressed, anxious, and it's affecting my sleep. I have been having trouble sleeping at night. I am also feeling lonely and blue. & Assistant: Lately, it's been pretty hard. I used to be the one everyone relied on when things got tough—helping with money or watching over the kids. But now, it's like they're all moving forward, and I'm just left behind, feeling like my efforts don't really matter. I can't shake the feeling of being stuck, like my life is at a standstill while everyone else's is moving on. It makes me question if all that sacrifice was worth anything. & Assistant: I can see how finding a supportive relationship might help, but it's a bit complicated for me right now. With my partner being away and the mixed feelings from friends about the situation, it’s hard to imagine starting something new. Plus, trusting someone new with all my baggage feels daunting. Maybe I need to work on connecting with people in other ways first. Do you think that's a good step? \\
        (see next page) & (see next page) & (see next page) \\
        \hline
    User: are you feeling depressed & User: do you want to speak to a counselor & User: yes, it will help certainly \\
    Assistant: Well, not really. But, I have had a lot of personal losses over the last few years. My dad died, my mom moved away, and my boyfriend is incarcerated. & Assistant: I've thought about speaking to a counselor, but I haven't made the move yet. It might be helpful to talk things through with someone who's unbiased and can provide some guidance. It feels like I could use a space where I can really explore what's going on with me without any judgment. Maybe it could help me sort through everything and find a bit more clarity. & Assistant: I'll try to keep that in mind. Building connections, even small ones, could be a step in the right direction. It's just been so long since I felt truly connected to someone outside of all this chaos. I hope taking small steps will help break down some of this loneliness. Thanks for your thoughts on this. \\
    \hline
    User: sorry to hear that & User: i can be of help with you & User: you are welcome \\
    Assistant: Thank you. It has been tough. I am feeling the lonliness the most right now. & Assistant: Lately, it feels like I'm trapped in this cycle where I keep helping my family, but it doesn't seem to matter much to them. I've been there for them financially and with the kids, but now they're all moving forward, and I'm just left behind. It's hard not to feel jealous, like I'm invisible in my own life. It's getting harder to shake off these feelings, and I can't help but wonder if I'll ever get out of this rut. & Assistant: I appreciate the support. It's nice to have a space where I can talk through these things, even if it's little by little. I hope these sessions help me find some clarity and strength to handle everything happening around me. Thank you for listening. \\
    \hline
        \dots & \dots & \dots \\
        \hline
    \end{tabular}
    \caption{Case Study of Expert Interactions with Three Chatbots}
    \label{tab:case2}
\end{table*}

\end{document}